\documentclass[11pt,a4paper]{article}
\usepackage[hyperref]{latell}
\usepackage{times}
\usepackage{latexsym}

\usepackage{microtype}
\usepackage{hyperref}
\usepackage{url}
\usepackage{booktabs}
\usepackage{amsmath}

\usepackage{lineno}
\usepackage{graphicx}
\usepackage{multirow}
\usepackage{booktabs}
\usepackage{ragged2e}
\usepackage{tabularx}
\usepackage{soul}

\usepackage{ifthen}   
\usepackage[dvipsnames]{xcolor} 
\usepackage{amssymb}  

\usepackage{tikz}        
\usepackage{xcolor}      
\usepackage{listings}    

\usepackage{booktabs}
\usepackage[table]{xcolor}
\usepackage{makecell}
\definecolor{lightgrayrow}{gray}{0.95}
\definecolor{headergray}{gray}{0.88}
\definecolor{subgroupgray}{gray}{0.975}

\usepackage{enumitem}
\usepackage{array}

\usepackage{dblfloatfix}

\newboolean{showcomments}
\setboolean{showcomments}{true}   

\ifthenelse{\boolean{showcomments}}{%
  \newcommand{\mynote}[2]{%
    \fbox{\bfseries\sffamily\scriptsize #1}%
    \ {\small$\blacktriangleright$ \textsf{\emph{#2}} $\blacktriangleleft$}%
  }%
}{%
  \newcommand{\mynote}[2]{}%
}

\usepackage{xcolor}
\usepackage{soul}
\usepackage{float}

\usepackage{xcolor}
\newcommand{\pos}[1]{\textcolor{green!60!black}{#1}}
\renewcommand{\neg}[1]{\textcolor{red!70!black}{#1}}

\soulregister\cite7
\soulregister\ref7
\soulregister\pageref7

\definecolor{MDYellow}{RGB}{255,243,176} 
\definecolor{MDGreen}{RGB}{209,236,196}  
\definecolor{MDOrange}{RGB}{255,214,165} 
\definecolor{MDBlue}{RGB}{198,221,255}   

\definecolor{darkblue}{rgb}{0, 0, 0.5}
\hypersetup{colorlinks=true, citecolor=darkblue, linkcolor=darkblue, urlcolor=darkblue}

\usepackage{booktabs}
\usepackage{multirow}
\usepackage[table]{xcolor}

\definecolor{HeatLow}{HTML}{F4F8FC}      
\definecolor{HeatHigh}{HTML}{8FB8DE}     
\definecolor{GainColor}{HTML}{2A9D8F}    
\definecolor{BaselineGray}{HTML}{F3F4F6} 

\newcommand{\heat}[2]{\cellcolor{HeatHigh!#1!HeatLow}{#2}}

\aclfinalcopy 
\def\aclpaperid{33} 

\title{LëtzCross: A Cross-Lingual Page-Level Benchmark for Multimodal Retrieval over Luxembourgish Documents}

\author{
\textbf{Omar El Bachyr\textsuperscript{1,*}}
\quad
\textbf{Fred Philippy\textsuperscript{1}}
\quad
\textbf{Laura Maria Bernardy\textsuperscript{1}}
\\[0.5em]
\textbf{Saad Ezzini\textsuperscript{2}}
\quad
\textbf{Jacques Klein\textsuperscript{1}}
\quad
\textbf{Tegawendé F. Bissyandé\textsuperscript{1}}
\\
\\
\textsuperscript{1}University of Luxembourg, Luxembourg \\
\textsuperscript{2}King Fahd University of Petroleum and Minerals, Saudi Arabia
\\
\small{
\textsuperscript{*}\textbf{Correspondence:}
\href{mailto:omar.elbachyr@uni.lu}{omar.elbachyr@uni.lu}
}
}

\date{}

\begin{document}
\maketitle
\begin{abstract}
Recent page-image retrievers such as ColPali~\citep{faysse2025colpali} have improved retrieval over visually rich documents, yet little is known about how they behave in cross-lingual, low-resource settings. We introduce \textit{LëtzCross}, a benchmark for cross-lingual page-level retrieval over Luxembourgish PDF documents, with document pages indexed as images and queries provided in English, French, German, and Luxembourgish. The benchmark combines text-focused QA pairs with visually grounded QA pairs, covering both textual and visual retrieval needs in PDF-based RAG. We use \textit{LëtzCross} to compare OCR-based text-only retrievers with ColPali-style page-image retrievers and find that the latter perform better across query languages in this system-level comparison. We also examine single-language and multilingual fine-tuning. Fine-tuning transfers across query languages, with French yielding the highest mean performance on Luxembourgish queries among the single-language settings. In the multilingual setting, including Luxembourgish gives the strongest results and substantially improves retrieval for Luxembourgish queries.
\end{abstract}

\section{Introduction}
Retrieval-augmented generation (RAG) and document retrieval systems increasingly operate in multilingual environments, where user queries and source documents may be written in different languages. In this setting, cross-lingual information retrieval (CLIR) has advanced substantially with dense retrieval architectures and multilingual encoders~\citep{karpukhin2020dense, conneau2020unsupervised}. However, most current evaluations focus on text-centric, high-resource scenarios and do not reflect the challenges of visually rich PDF documents.

Recent vision--language retrieval models, such as DSE~\citep{ma2024unifying} and ColPali~\citep{faysse2025colpali}, improve retrieval over document pages by jointly modeling textual and visual evidence. Yet their behavior in low-resource language settings remains underexplored. We study this challenge through the case of Luxembourgish, a low-resource language for which NLP resources are growing but still limited compared to high-resource languages~\citep{luxembert, luxgpt, luxt5}.

To address this gap, we introduce \textit{LëtzCross}, a benchmark for cross-lingual page-level retrieval over Luxembourgish PDF pages, where pages are represented as images. The benchmark combines text-focused and visually grounded QA pairs, and is designed around realistic page-level retrieval needs in RAG pipelines, where answers may depend on textual content, visual structure, or both, and where queries may be issued in multiple languages.

Our main contributions are:
\begin{itemize}[itemsep=1pt, topsep=2.6pt]
    \item We introduce LëtzCross, a benchmark for cross-lingual page-level retrieval over Luxembourgish PDF pages, where pages are represented as images.
    \item We benchmark multilingual text-only retrievers and late-interaction page-image retrievers on this task.
    \item We study query-language-specific fine-tuning for cross-lingual retrieval over Luxembourgish document pages.
    \item We analyze multilingual fine-tuning settings for retrieval over Luxembourgish document pages as a low-resource language use case.
\end{itemize}

Overall, our findings show that late-interaction page-image retrievers are strong candidates for cross-lingual retrieval over Luxembourgish document pages, outperforming the evaluated OCR-based text-only baselines in a system-level comparison. We further show that query-language-specific fine-tuning transfers across languages, while multilingual fine-tuning is most effective when it includes the target low-resource language. Together, these results position \textit{LëtzCross} as a useful benchmark for studying cross-lingual page-level retrieval over Luxembourgish PDF documents as a low-resource language in RAG settings.

We release LëtzCross and all experimental code and resources through our \href{https://github.com/OmarElbachyr/letzcross-benchmark}{GitHub repository}.

\section{Related Work}
\paragraph{Luxembourgish NLP-Resources}
The Luxembourgish NLP landscape is still under development. Natural Language Processing techniques for Luxembourgish are primarily developed within a low-resource setting and rely heavily on manual efforts, such as data collection and annotation. In recent years, this approach has enabled the development of several foundational resources and applications, including sentiment analysis systems \citep{sirajzade, gierschek}, a Luxembourgish dependency parsing treebank \citep{plumdepend}, and zero-shot topic classification \citep{philippy}. These initiatives have contributed to a better understanding of the linguistic properties of Luxembourgish through computational methods.
Within this context, Lothritz et al. introduced LuxemBERT \citep{luxembert}, a language model trained using artificial data augmentation strategies. In addition to the model itself, they expanded Luxembourgish resources for several downstream tasks, including part-of-speech tagging, named entity recognition, and news classification. Two further language models have since been established for Luxembourgish: LuXGPT \citep{luxgpt}, developed using transfer learning techniques, and, more recently, LUXT5 \citep{luxt5}, which benefits from multilingual pretraining.
Beyond language modeling, practical NLP applications have also emerged. These include LUX-ASR \citep{gilles2} for automatic speech recognition, as well as systems for automatic comment moderation \citep{ranasinghe} and orthographic correction \citep{purschke}.

\paragraph{Cross-lingual retrieval}
Transformer-based Cross-Lingual Information Retrieval (CLIR) builds on multilingual representation learning and dense retrieval architectures that enable semantic matching across languages. Early foundations include BERT~\citep{devlin2019bert}, which introduced contextual transformer representations for retrieval models, and XLM-R~\citep{conneau2020unsupervised}, a multilingual transformer that provides strong cross-lingual embeddings. Dense Passage Retrieval (DPR)~\citep{karpukhin2020dense} later established the dual-encoder architecture widely used in dense retrieval systems. Building on these foundations, more recent work explores unsupervised and generative approaches, such as Unsupervised Multilingual Dense Retrieval (UMR) proposed by~\citep{huang2024unsupervised}, which uses generative pseudo-labeling to train dense retrievers without parallel supervision, and research on adapting LLMs for retrieval tasks~\citep{guo2024steering}. Recent multilingual embedding models such as BGE-M3~\citep{bge-m3}, Qwen3 Embedding~\citep{qwen3embedding}, and voyage-4-nano~\citep{voyage4nano2026} further improve cross-lingual semantic representations and are widely used as baselines in modern CLIR systems. 

\paragraph{Vision Language Retrieval}
Text-only encoders have achieved strong performance in information retrieval (IR), with models such as \citep{bge-m3, qwen3embedding, voyage4nano2026} demonstrating high effectiveness across multilingual and dense retrieval tasks. 
However, recent studies in IR have advanced beyond conventional text-only systems by incorporating visual features. DSE~\citep{ma2024unifying} propose the use of Vision–Language Models (VLMs) to encode full document screenshots, successfully preserving both text and visual information without the need for optical character recognition (OCR). Similarly, ColPali~\citep{faysse2025colpali} addresses the shortcomings of text-focused retrieval in the context of visually rich documents by adapting ColBERT~\citep{khattab2020colbert} late-interaction similarity scoring mechanism to work with page images. This produces multi-vector visual embeddings that enable fine-grained matching between query tokens and specific visual areas on document pages.
\begin{figure*}[ht]
  \centering
  \scalebox{1}{%
    \includegraphics[width=\linewidth]{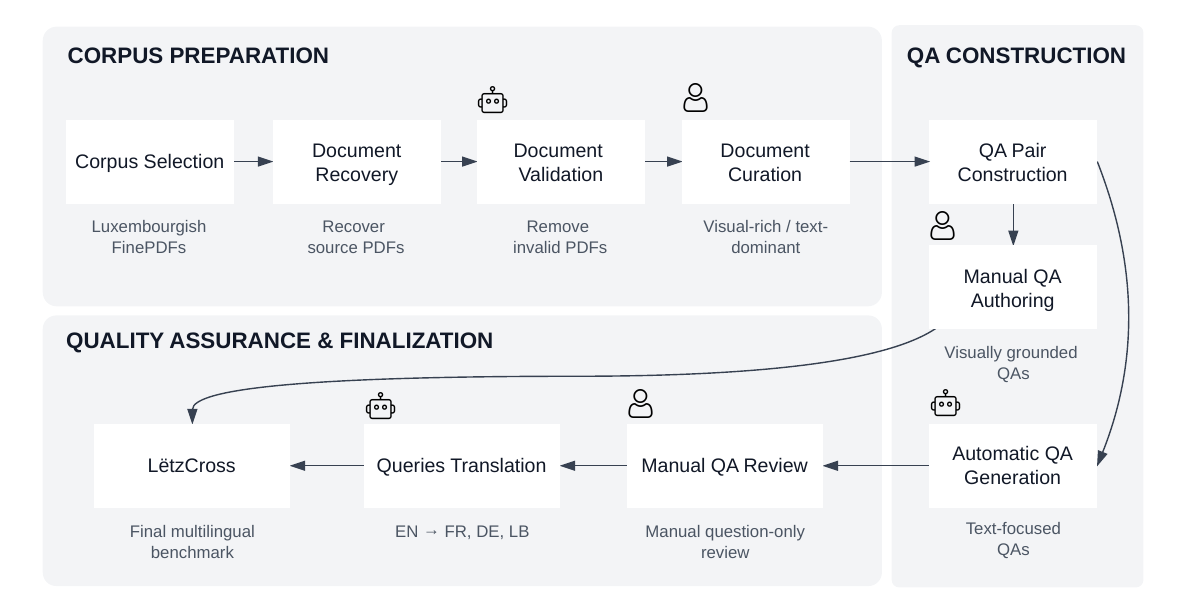}
  }
  \caption{Overview of the LëtzCross benchmark construction pipeline.}
  \label{fig:benchmark_pipeline}
\end{figure*}

\section{LëtzCross Benchmark Construction}
\textit{LëtzCross} is a cross-lingual page-level retrieval benchmark for Question-Answering (QA), built from Luxembourgish documents, with queries in English (EN), French (FR), German (DE), and Luxembourgish (LB). Documents are represented as page images, and answers are grounded in the corresponding pages. Figure~\ref{fig:benchmark_pipeline} summarizes the process, with steps described below.

\subsection{PDF Documents Acquisition}
We obtain the original PDF documents by iterating over the Luxembourgish (LB) split of FinePDFs~\citep{kydlicek2025finepdfs} dataset and retrieving each document from its source. When documents are archived in Common Crawl, we extract the corresponding PDF directly from the web archive; otherwise, we download the PDF from the original URL provided in the dataset. This process allows us to recover the original PDF files associated with the dataset entries.

\subsection{Filtering and Curation of PDF Documents}
After downloading the PDFs, we perform an automatic validation step to remove corrupted or unreadable documents by verifying that each PDF can be successfully parsed and processed. We then manually classify the validated PDFs into visually rich and text-dominant subsets. A document is considered visually rich when it contains informative visual elements, such as charts, diagrams, or tables, that convey substantive information beyond the surrounding text. Non-informative visuals, such as portraits, landscapes, and logos, are disregarded. Documents whose information is conveyed primarily through continuous text, with few or no informative visual elements, are classified as text-dominant. The resulting corpus serves as the foundation for the subsequent Question--Answer generation step.
\subsection{Question–Answer Generation}
We then use the validated PDFs from the previous step to generate Question–Answer (QA) pairs. We generate queries in English to ensure high-quality and consistent question formulation, as this yields more reliable QA pairs on Luxembourgish pages. The benchmark combines two complementary QA sources: manually authored visually grounded QA pairs for visually rich pages, and automatically generated text-focused QA pairs for text-dominant pages.

\noindent\paragraph{Manual visually grounded QA authoring.}
For the subset of PDFs containing rich visual components, a native Luxembourgish speaker manually crafts page-level QA pairs that intentionally target visually grounded evidence, such as charts, plots, diagrams, figures, and other layout-dependent elements, rather than relying solely on textual content. Questions are phrased to be self-contained and answerable from a single page image, with concise answers grounded in the visual content.

\noindent\paragraph{Automatic text-focused QA generation.}
\label{sec:auto_qa_generation}
Using the text-dominant PDF subset, we automatically generate page-level QA pairs with GPT-5-mini. Each page is processed independently as an image and prompted to produce self-contained questions with concise answers grounded in its textual content. To validate the generated QA pairs, each query and its corresponding PDF are provided to Gemini-2.5-Flash, which extracts an answer from the document. We then compare the extracted answer with the \textit{gold answer} using both exact and semantic matching; the latter is performed by prompting GPT-5-mini to assess whether the two answers convey the same meaning. Overall, 90.28\% of the QA pairs were judged semantically equivalent by GPT-5-mini, and 71.50\% contained the exact literal gold answer. QA pairs that do not yield consistent answers are discarded, resulting in a consistency-filtered set for the subsequent manual validation step.

\subsection{Quality Control and RAG Alignment}
As a follow-up step, the manually authored QA pairs are included directly in the final benchmark. In contrast, the automatically generated QA pairs undergo a manual validation phase focused exclusively on question formulation: annotators are shown only the questions and revise them when needed to produce clear, natural queries that reflect RAG-relevant information needs (e.g., realistic, retrieval-oriented queries). 

\subsection{Cross-Lingual Query Translation}
\label{sec:qa_translation}
To enable cross-lingual retrieval evaluation, we translate the finalized English queries into German, French, and Luxembourgish using a controlled LLM prompt that enforces semantic preservation and prevents paraphrasing or additional explanations. This produces aligned query sets across languages, where each translated query is intended to preserve the same retrieval intent as the original English query. We use Gemini-3-Flash for German and French, and Gemini-3-Pro for Luxembourgish. This choice was made after manually inspecting a small set of trial translations: Gemini-3-Pro produced better Luxembourgish translations, while Gemini-3-Flash was sufficient for German and French.

The final benchmark contains 579 QA pairs over 908 document pages. Table~\ref{tab:benchmark_stats} summarizes the benchmark statistics, including the distribution between automatically generated text-focused QA pairs and manually authored visually grounded QA pairs. All prompts used during benchmark construction are available in the accompanying repository.

\begin{table}[t]
\centering
\footnotesize
\setlength{\tabcolsep}{4pt}
\renewcommand{\arraystretch}{0.95}
\begin{tabular}{@{}lr@{}}
\toprule
\textbf{Benchmark statistic} & \textbf{Value} \\
\midrule
Number of pages & 908 \\
Number of queries & 579 \\
Avg. query length & 19.5 \\
Avg. corpus item length & 722.5 \\
\midrule
\multicolumn{2}{@{}l}{\textbf{QA type distribution}} \\
Auto-generated text-focused & 530 \\
Manual visually grounded & 49 \\
\bottomrule
\end{tabular}
\caption{Overview statistics of the \textit{LëtzCross} evaluation benchmark. Query and corpus item lengths are reported in tokens.} 
\label{tab:benchmark_stats}
\end{table}

\section{Methods}
This section presents our method for investigating cross-lingual retrieval over Luxembourgish document pages as a low-resource case study using late-interaction page-image retrievers.

\subsection{Training Data Construction}
We construct the training data by scraping Luxembourgish Wikipedia articles and converting them into PDF documents. We first apply size-based filtering with minimum and maximum page thresholds to exclude trivial or overly long documents (e.g., pages containing only footnotes). QA pairs are then generated from 4,977 PDFs with 11,833 pages using the same automatic text-focused QA generation procedure as in the benchmark (Section~\ref{sec:auto_qa_generation}). This resulted in 22,028 QA pairs. In addition, we construct 450 QA pairs from the same Wikipedia-derived pipeline and reserve them as a held-out development split for fine-tuning.
To extend the training set for cross-lingual retrieval, we translate queries into German and French, and a 5,000-query subset into Luxembourgish, following the same query translation procedure as in the benchmark (Section~\ref{sec:qa_translation}). 


\subsection{Retriever and Training Objective}
We fine-tune a late-interaction page-image retriever based on ColPali~\citep{faysse2025colpali} using the training setup provided by the underlying framework. Each document page is treated as an image and split into visual patches, which are encoded into contextualized image patch embeddings, while the query text is tokenized and encoded into query token embeddings. Both are represented in a shared latent space, and relevance is computed through late interaction between query tokens and page patches.

Let $q_i = \{q_{i1}, \dots, q_{i|q_i|}\}$ denote the query token embeddings of query $i$, and let $d_j = \{d_{j1}, \dots, d_{j|d_j|}\}$ denote the document image patch embeddings of page $j$. The score between query $i$ and page $j$ is:
\begin{equation}
\text{score}(q_i, d_j)
= \sum_{t=1}^{|q_i|}
\max_{u=1,\dots,|d_j|}
\left(q_{it}^{\top} d_{ju}\right)  
\label{eq:late_interaction_score}
\end{equation}

We use the original ColBERT~\citep{khattab2020colbert} contrastive objective with in-batch negatives:
\begin{equation}
\mathcal{L}
= -\frac{1}{B}
\sum_{i=1}^{B}
\log
\frac{
\exp\left(\text{score}(q_i, d_i)/\tau\right)
}{
\sum_{j=1}^{B}
\exp\left(\text{score}(q_i, d_j)/\tau\right)
}
\label{eq:contrastive_loss}
\end{equation}

where $\tau$ is a temperature hyperparameter. Each training instance consists of a query paired with its corresponding document page image as the positive example, while the remaining $B-1$ pages in the batch serve as in-batch negatives. We do not use external hard negatives or additional negative sampling.

\subsection{Fine-Tuning Configurations}

We consider two fine-tuning settings. In the \textit{per-language} setting, the retriever is fine-tuned separately on English, French, and German queries. In the \textit{multilingual} setting, a single model is trained on mixed-language data, using either EN+FR+DE or EN+FR+DE+LB. In all cases, document pages remain in Luxembourgish, and only the query language varies. To analyze visual adaptation, we also compare the default setting, which updates both textual and visual components, with a text-component-only variant. The EN, FR, and DE settings each use 22,028/450 train/dev QA pairs, while EN+FR+DE and EN+FR+DE+LB use 21,000/1,350 and 20,000/1,800 train/dev samples, respectively.
\begin{table*}[t]
\centering
\scalebox{0.82}{%
\begin{tabular}{
    l
    *{4}{>{\centering\arraybackslash}p{0.62cm}}
    ccc
}
\toprule
\textbf{Retriever} &
\multicolumn{4}{c}{\textbf{Query Language}} &
\textbf{Avg.} &
\textbf{Index (ms/page) $\downarrow$} &
\textbf{Latency (ms/query) $\downarrow$} \\
\cmidrule(lr){2-5}
& \textbf{EN} & \textbf{FR} & \textbf{DE} & \textbf{LB} & & & \\
\midrule

\rowcolor{lightgrayrow}
\multicolumn{8}{l}{\textsc{text-only retrievers}} \\

bge-m3
& 72.48 & 71.63 & 72.47 & 76.12 & 73.18
& $21.714 \pm 0.015$ & $3.859 \pm 0.027$ \\

Qwen3-Embedding-0.6B
& 73.02 & 71.63 & 75.21 & 67.97 & 71.96
& $42.863 \pm 0.115$ & $6.935 \pm 0.008$ \\

multilingual-e5-large
& 75.96 & 75.02 & 77.24 & 68.32 & 74.14
& $11.068 \pm 0.070$ & $3.858 \pm 0.012$ \\

jina-embeddings-v4
& 76.51 & 75.39 & 77.30 & 73.86 & 75.76
& $113.007 \pm 0.287$ & $26.271 \pm 0.336$ \\

\midrule

\rowcolor{lightgrayrow}
\multicolumn{8}{l}{\textsc{multi-vector page-image retrievers}} \\

colSmol-256M
& 71.61 & 59.45 & 52.76 & 55.11 & 59.73
& $367.108 \pm 1.415$ & $3.914 \pm 0.035$ \\

colSmol-500M
& 73.90 & 67.56 & 65.92 & 63.87 & 67.81
& $361.103 \pm 11.169$ & $4.125 \pm 0.017$ \\

colpali-v1.3
& 77.03 & 77.43 & 78.46 & 75.78 & 77.18
& $61.801 \pm 0.110$ & $2.997 \pm 0.007$ \\

colqwen2.5-v0.2
& \underline{78.07} & \textbf{78.71} & \textbf{79.92}
& \textbf{78.60} & \textbf{78.83}
& $133.876 \pm 0.813$ & $4.892 \pm 0.113$ \\

colnomic-embed-multimodal-3b
& \textbf{78.73} & \underline{78.44} & \underline{79.59}
& \underline{77.89} & \underline{78.66}
& $145.193 \pm 0.277$ & $4.981 \pm 0.067$ \\

\bottomrule
\end{tabular}%
}
\caption{Cross-lingual retrieval over Luxembourgish document pages. Query-language columns report nDCG@10 where best results are shown in \textbf{bold}, and second-best results are \underline{underlined}. Efficiency values are mean $\pm$ standard deviation over three runs on the same hardware.}
\label{tab:benchmark_results}
\end{table*}

\section{Experimental Setup}
\paragraph{Models.}
Based on the input type, we consider two families of retrievers in our experiments: text-only retrievers and late-interaction page-image retrievers. The text-only baselines are bge-m3~\citep{bge-m3}, Qwen3-Embedding-0.6B~\citep{qwen3embedding}, multilingual-e5-large~\citep{wang2024multilingual}, and jina-embeddings-v4~\citep{günther2025jinaembeddingsv4universalembeddingsmultimodal}\footnote{Although \texttt{jina-embeddings-v4} is a multimodal embedding model, we use it here in a text-only setting.}. The late-interaction page-image retrievers are colSmol-256M, colSmol-500M, colpali-v1.3, and colqwen2.5-v0.2~\citep{faysse2025colpali}, and colnomic-embed-multimodal-3b~\citep{nomic_colnomic_embed_multimodal_3b}.

\paragraph{Text Extraction Pipeline.}
\label{sec:ocr_pipeline}
The text-only retrievers described above use page-level text extracted with \texttt{unstructured[pdf]} v0.18.15 under the \texttt{hi\_res} strategy, combining Poppler page rendering, YOLOX layout detection, Luxembourgish Tesseract OCR (\texttt{ltz}), and table-structure extraction. Existing PDF text was incorporated when available, and all PDFs were processed using the same pipeline. We consider this a practical system-level comparison between text-based and page-image retrieval, rather than an evaluation of PDF extraction methods.

\paragraph{Fine-Tuning Details.}
We fine-tune the retriever using the ColBERT late-interaction contrastive loss defined in Equation~\eqref{eq:contrastive_loss} with temperature $\tau=0.02$ and LoRA adapters (\texttt{r}=32, \texttt{alpha}=32, dropout $=0.1$). In the text-component-only setting, LoRA targets the query encoder's attention projections (\texttt{q\_proj}, \texttt{k\_proj}, \texttt{v\_proj}, and \texttt{o\_proj}), feed-forward projections (\texttt{down\_proj}, \texttt{gate\_proj}, and \texttt{up\_proj}), and \texttt{custom\_text\_proj}; the vision encoder remains frozen. In the text+visual setting, the same attention and feed-forward projections are adapted in both the textual and visual transformer blocks, where the visual component denotes the blocks encoding page-image patches. All untargeted parameters remain frozen. Training is performed for 3 epochs with a learning rate of $5\times10^{-5}$ and 100 warmup steps, using a per-device batch size of 16 on 4 NVIDIA L40S GPUs and no gradient accumulation, resulting in an effective batch size of 64. All runs use bfloat16 precision. 
Each fine-tuning configuration is repeated with three seeds (42, 43, and 44),
and results are reported as mean $\pm$ standard deviation.

\paragraph{Evaluation Setup and Metrics.}
We evaluate retrieval at the page level: for each query, the retriever ranks all document pages in the benchmark against the gold relevant page. Results are reported separately for English, French, German, and Luxembourgish queries. Our main metric is nDCG@10; for analyses across retrieval depths, we additionally report nDCG@$k$ for $k \in \{1,3,5,10\}$. Indexing time per page and query latency per query are aggregated across the four query languages and reported as mean $\pm$ standard deviation over three runs. All evaluation runs, including effectiveness and efficiency measurements, use one NVIDIA L40S GPU and eight CPU cores.





\section{Results and Analysis}
Our experiments evaluate cross-lingual retrieval over Luxembourgish document pages. We first compare text-only and page-image retrievers, then analyze query-language-specific fine-tuning, including an ablation on the visual component, and finally evaluate multilingual fine-tuning. 

\subsection{Page-Image vs. OCR-Based Text Retrieval over Luxembourgish Documents}

We compare multilingual OCR-based text-only embedding baselines, which index
page-level text produced by the OCR, layout-analysis, and table-extraction
pipeline described in Section~\ref{sec:ocr_pipeline}, with late-interaction page-image retrievers ranging from lightweight models below 500M parameters to models of up to 3B parameters. Table~\ref{tab:benchmark_results} shows that the strongest page-image retrievers outperform the evaluated text-only baselines across all query languages. Since the two families differ in input representation, extraction pipeline, and model architecture, we interpret this as a practical system-level comparison rather than a controlled modality ablation.

Performance remains strongly model-dependent. The smaller colSmol models underperform strong multilingual text-only retrievers such as jina-embeddings-v4. In contrast, colqwen2.5-v0.2 and colnomic-embed-multimodal-3b outperform all evaluated text-only models across every query language, while colpali-v1.3 also performs strongly on English, French, and German. This shows that sufficiently capable multi-vector page-image retrievers can outperform multilingual text embedding models in cross-lingual retrieval. The best page-image retriever achieves an average nDCG@10 of 78.83, improving by 3.07 points over the strongest text-only baseline. A more customized OCR and layout-detection pipeline could nevertheless affect this gap.

These results are notable because the text-only baselines are explicitly designed for multilingual retrieval, whereas the page-image retrievers primarily target visual-document retrieval. Their strong zero-shot cross-lingual performance may partly reflect the multilingual VLM backbones used by several models.
\begin{table*}[t]
\centering
\scalebox{0.9}{%
\begin{tabular}{lccccc}
\toprule
\multirow{2}{*}{\textbf{Training setup}} & \multicolumn{5}{c}{\textbf{Query language}} \\
\cmidrule(lr){2-6}
 & \textbf{EN} & \textbf{FR} & \textbf{DE} & \textbf{LB} & \textbf{Avg.} \\
\midrule
\rowcolor{BaselineGray}
Baseline
& 78.73
& 78.44
& 79.59
& 77.89
& 78.66 \\
\midrule
FT (EN)
& \heat{96}{\textbf{81.66} {\scriptsize $\pm$ 0.33 \pos{(+2.93)}}}
& \heat{94}{81.34 {\scriptsize $\pm$ 0.40 \pos{(+2.89)}}}
& \heat{73}{\textbf{81.84} {\scriptsize $\pm$ 0.11 \pos{(+2.24)}}}
& \heat{20}{78.48 {\scriptsize $\pm$ 0.65 \pos{(+0.60)}}}
& 80.83 {\scriptsize $\pm$ 0.23 \pos{(+2.17)}} \\
FT (FR)
& \heat{78}{81.11 {\scriptsize $\pm$ 0.57 \pos{(+2.38)}}}
& \heat{100}{\textbf{81.50} {\scriptsize $\pm$ 0.49 \pos{(+3.06)}}}
& \heat{71}{81.76 {\scriptsize $\pm$ 0.40 \pos{(+2.17)}}}
& \heat{51}{\textbf{79.45} {\scriptsize $\pm$ 0.75 \pos{(+1.57)}}}
& \textbf{80.96} {\scriptsize $\pm$ 0.55 \pos{(+2.29)}} \\
FT (DE)
& \heat{25}{79.48 {\scriptsize $\pm$ 0.27 \pos{(+0.75)}}}
& \heat{68}{80.51 {\scriptsize $\pm$ 0.88 \pos{(+2.07)}}}
& \heat{64}{81.56 {\scriptsize $\pm$ 0.34 \pos{(+1.97)}}}
& \heat{47}{79.34 {\scriptsize $\pm$ 0.33 \pos{(+1.45)}}}
& 80.22 {\scriptsize $\pm$ 0.14 \pos{(+1.56)}} \\
\midrule
\textbf{Avg. FT}
& 80.75 {\scriptsize $\pm$ 0.15}
& 81.12 {\scriptsize $\pm$ 0.22}
& 81.72 {\scriptsize $\pm$ 0.08}
& 79.09 {\scriptsize $\pm$ 0.34}
& 80.67 {\scriptsize $\pm$ 0.11} \\
\bottomrule
\end{tabular}
}
\caption{nDCG@10 by query language (columns) for the baseline model and models fine-tuned on different training languages (rows). Values are reported as mean $\pm$ standard deviation over three runs; values in parentheses indicate gains over the baseline on the same query language. Best values are shown in \textbf{bold}.}
\label{tab:transfer_ndcg10}
\end{table*}

Table~\ref{tab:qa_source_split} further examines performance by QA source. The page-image retrievers outperform the strongest text-only baseline on both the text-focused and visually grounded subsets, with larger gains on the visually grounded questions. This provides additional evidence that page-image retrieval is beneficial beyond the predominantly text-focused portion of the benchmark. However, the visually grounded subset contains only 49 QA pairs and yields higher scores for all retrievers, so these results remain descriptive and should be interpreted cautiously.

\begin{table}[t]
\centering
\small
\setlength{\tabcolsep}{4pt}
\resizebox{\columnwidth}{!}{%
\begin{tabular}{@{}lccc@{}}
\toprule
\textbf{Retriever} & \textbf{Text-focused} & \textbf{Visually grounded} & \textbf{Overall} \\
\midrule
jina-embeddings-v4 & 75.27 & 81.10 & 75.76 \\
colqwen2.5-v0.2 & 78.42 & 83.23 & 78.83 \\
colnomic-embed-multimodal-3b & 77.91 & 86.82 & 78.66 \\
\bottomrule
\end{tabular}%
}
\caption{Descriptive nDCG@10 split by QA source, averaged across query languages.}
\label{tab:qa_source_split}
\end{table}

\paragraph{Efficiency.}
Table~\ref{tab:benchmark_results} shows clear efficiency differences across retrievers. Among text-only models, multilingual-e5-large indexes fastest, while jina-embeddings-v4 is slower, likely due to its larger multimodal backbone. Among page-image retrievers, colpali-v1.3 is the fastest, possibly because of its fixed-resolution processing and fixed visual-token budget. The slower ColSmol indexing further shows that runtime depends on visual-token count and implementation choices, not only parameter count. Query latency remains similar across most page-image models because page representations are precomputed.

For the text-only retrievers, page parsing introduces an additional shared preprocessing cost. Using the extraction pipeline described in Section~\ref{sec:ocr_pipeline}, parsing takes $5.382 \pm 0.054$ s/page across three runs. Including this cost, the end-to-end preprocessing and indexing time ranges from $5.393 \pm 0.054$ s/page for multilingual-e5-large to $5.495 \pm 0.054$ s/page for jina-embeddings-v4. Thus, for text-only retrieval, document parsing dominates the overall offline preprocessing cost.

\subsection{Cross-Lingual Transfer of Query-Language-Specific Fine-Tuning}

This subsection studies how fine-tuning transfers across evaluation languages for the late-interaction page-image retrievers considered in this work. We fine-tune the retriever separately on English, French, and German queries, and evaluate each variant on all four query languages. We use colnomic-embed-multimodal-3b for this analysis, as both the baseline model and the backbone for fine-tuning, because its performance remains very close to the top model in our experiments, while ViDoRe leaderboard~\citep{loison2026vidorev3comprehensiveevaluation} results suggest it is a strong page-image retriever overall.

Table~\ref{tab:transfer_ndcg10} shows that all language-specific fine-tuning settings improve over the baseline on every query language. FT (FR) achieves the highest overall average at 80.96, closely followed by FT (EN) at 80.83, while FT (DE) reaches 80.22. The largest language-specific gains are obtained by FT (EN) on English (+2.93) and FT (FR) on French (+3.06). Cross-lingual transfer is also substantial: FT (EN) achieves the highest German score at 81.84 (+2.24), while FT (FR) performs best on Luxembourgish at 79.45 (+1.57).

These results show that the strongest configuration does not always correspond to matching the fine-tuning and evaluation languages. In particular, English fine-tuning transfers strongly to German, while French fine-tuning provides the strongest Luxembourgish performance. German fine-tuning improves all four languages but produces the lowest overall average among the three variants. Overall, query-language-specific fine-tuning consistently improves multilingual retrieval, while the magnitude of transfer depends on the training and evaluation language pair.

\begin{figure*}[t]
  \centering
  \scalebox{1}{%
    \includegraphics[width=\textwidth]{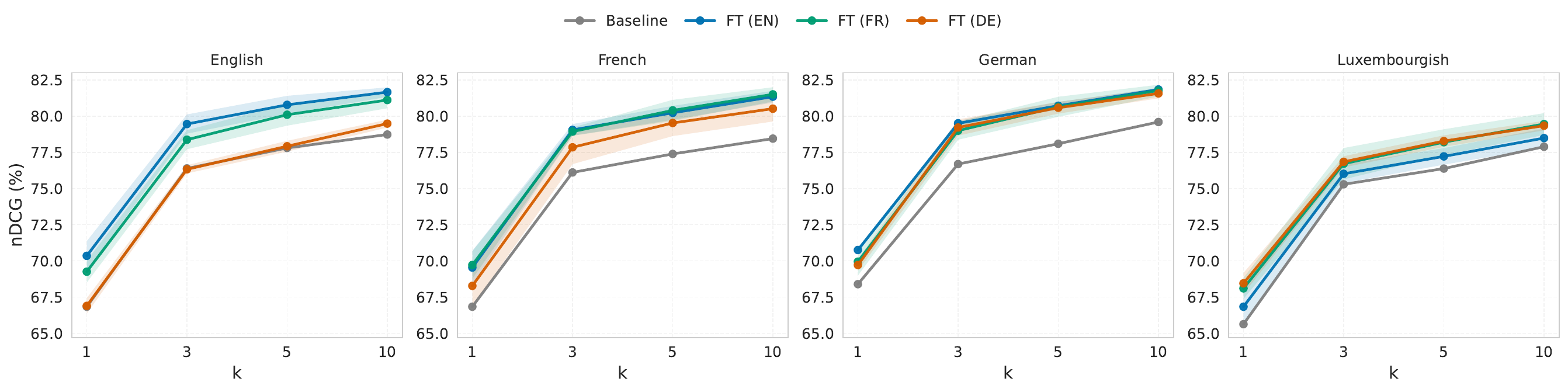}
  }
 \caption{Multilingual retrieval performance (nDCG@$k$) across query languages. Shaded bands indicate $\pm 1$ standard deviation across three fine-tuning runs.}
  \label{fig:per_lang_finetunes_comparison}
\end{figure*}

Figure~\ref{fig:per_lang_finetunes_comparison} shows that these trends are broadly maintained across retrieval cutoffs. FT (EN) remains strongest for English queries, whereas the fine-tuned variants are much closer for French and German. For Luxembourgish, FT (FR) and FT (DE) generally perform better than FT (EN). The uncertainty bands also overlap in several cases, indicating that small differences between fine-tuning languages should not be over-interpreted. Overall, the improvements persist across different values of $k$, showing that the cross-lingual benefits of fine-tuning are not limited to nDCG@10.


\paragraph{Effect of Updating the Visual Component.}
In all previous fine-tuning experiments, we use a setting in which both the textual and visual components of the page-image retriever are updated. To isolate the contribution of visual adaptation, we compare this setting with the corresponding text-component-only fine-tuning variant. Table~\ref{tab:visual_ablation} reports the paired difference in nDCG@10 between the two settings across three runs.

The mean differences are positive for every training and evaluation language combination, indicating that updating the visual component provides an additional benefit beyond text-component-only adaptation. The magnitude of this gain varies substantially across settings, ranging from 0.21 points for FT (EN) evaluated on Luxembourgish to 2.53 points for FT (DE) evaluated on French. The larger gains observed for some FT (DE) configurations are accompanied by higher run-to-run variability, while several EN and FR configurations show smaller but more stable improvements. Overall, these results suggest that visual adaptation provides complementary gains, although its contribution depends on both the fine-tuning language and the evaluation language.
\begin{table}[t]
\small
\centering
\scalebox{0.8}{%
\begin{tabular}{lcccc}
\toprule
\multirow{2}{*}{\textbf{Training setup}} & \multicolumn{4}{c}{\textbf{Query language}} \\
\cmidrule(lr){2-5}
 & \textbf{EN} & \textbf{FR} & \textbf{DE} & \textbf{LB} \\
\midrule
FT (EN)
& \heat{67}{1.69 {\scriptsize $\pm$ 0.57}}
& \heat{59}{1.49 {\scriptsize $\pm$ 0.53}}
& \heat{56}{1.41 {\scriptsize $\pm$ 0.12}}
& \heat{8}{0.21 {\scriptsize $\pm$ 0.38}} \\
FT (FR)
& \heat{90}{2.27 {\scriptsize $\pm$ 0.16}}
& \heat{52}{1.32 {\scriptsize $\pm$ 0.34}}
& \heat{70}{\textbf{1.78} {\scriptsize $\pm$ 0.44}}
& \heat{50}{\textbf{1.27} {\scriptsize $\pm$ 0.98}} \\
FT (DE)
& \heat{98}{\textbf{2.49} {\scriptsize $\pm$ 1.25}}
& \heat{100}{\textbf{2.53} {\scriptsize $\pm$ 1.46}}
& \heat{53}{1.34 {\scriptsize $\pm$ 1.07}}
& \heat{32}{0.82 {\scriptsize $\pm$ 0.48}} \\
\bottomrule
\end{tabular}
}
\caption{Difference in nDCG@10 between text+visual and text-component-only fine-tuning across query languages. Values are reported as mean $\pm$ standard deviation over three runs; positive values indicate gains from updating the visual component. Best values are shown in \textbf{bold}.}
\label{tab:visual_ablation}
\end{table}

\begin{table*}[t]
\centering
\scalebox{0.8}{%
\begin{tabular}{lccccc}
\toprule
\multirow{2}{*}{\textbf{Training setup}} & \multicolumn{5}{c}{\textbf{Query language}} \\
\cmidrule(lr){2-6}
 & \textbf{EN} & \textbf{FR} & \textbf{DE} & \textbf{LB} & \textbf{Avg.} \\
\midrule

Baseline
& 78.73
& 78.44
& 79.59
& 77.89
& 78.66 \\
\midrule

\rowcolor{lightgrayrow}
FT (FR)
& \textbf{81.11 $\pm$ 0.57}
& \textbf{81.50 $\pm$ 0.49}
& \textbf{81.76 $\pm$ 0.40}
& \underline{79.45 $\pm$ 0.75}
& \textbf{80.96 $\pm$ 0.55} \\

\midrule
\multicolumn{6}{l}{\textbf{Multilingual FT}} \\

EN+FR+DE
& 80.47 $\pm$ 0.39 {\scriptsize \neg{(-0.63)}}
& 80.46 $\pm$ 0.16 {\scriptsize \neg{(-1.04)}}
& 81.30 $\pm$ 0.26 {\scriptsize \neg{(-0.46)}}
& 77.48 $\pm$ 0.48 {\scriptsize \neg{(-1.97)}}
& 79.93 $\pm$ 0.12 {\scriptsize \neg{(-1.03)}} \\

EN+FR+DE+LB
& \underline{81.03 $\pm$ 0.51} {\scriptsize \neg{(-0.08)}}
& \underline{81.04 $\pm$ 0.44} {\scriptsize \neg{(-0.46)}}
& \underline{81.66 $\pm$ 0.44} {\scriptsize \neg{(-0.10)}}
& \textbf{80.02 $\pm$ 0.56} {\scriptsize \pos{(+0.57)}}
& \underline{80.94 $\pm$ 0.41} {\scriptsize \neg{(-0.02)}} \\

\bottomrule
\end{tabular}
}
\caption{nDCG@10 across query languages for the baseline, the best single-language fine-tuning setting, and multilingual fine-tuning variants. Results are reported as mean $\pm$ standard deviation over three runs. Values in parentheses indicate changes relative to FT (FR); the best value is shown in \textbf{bold} and the second-best is \underline{underlined}.}
\label{tab:multilingual_ft_results}
\end{table*}
\subsection{Multilingual Fine-Tuning for Improved Cross-Lingual Retrieval}

We further investigate whether multilingual fine-tuning improves cross-lingual retrieval beyond the single-language fine-tuning studied earlier. We compare the best single-language setting, FT (FR), with two multilingual variants: one fine-tuned on English, French, and German, and another fine-tuned on English, French, German, and Luxembourgish. We again use colnomic-embed-multimodal-3b as both the baseline model and the backbone used for fine-tuning.

As shown in Table~\ref{tab:multilingual_ft_results}, multilingual fine-tuning on English, French, and German underperforms FT (FR) across all query languages, with an average decrease of 1.03 nDCG points. The largest decrease occurs on Luxembourgish queries ($-1.97$). This degradation may partly reflect the smaller number of training examples available per language in the multilingual setting compared with single-language FT (FR), which can reduce language-specific adaptation. The absence of Luxembourgish supervision may further contribute to the larger drop observed on Luxembourgish queries.

Adding Luxembourgish substantially changes this pattern. The EN+FR+DE+LB setting reaches an average nDCG@10 of 80.94, essentially matching FT (FR) at 80.96. Relative to EN+FR+DE, adding Luxembourgish improves performance across all query languages, with the largest increase on Luxembourgish queries, from 77.48 to 80.02 ($+2.54$ points). Compared with FT (FR), the multilingual setting remains close on English, French, and German, with differences of $-0.08$, $-0.46$, and $-0.10$, respectively, while improving Luxembourgish retrieval by $+0.57$.

Since the corpus pages are in Luxembourgish, direct exposure to Luxembourgish queries may help align query representations with the corresponding document pages. However, the multilingual settings also differ from single-language fine-tuning in the amount and distribution of training data per language, so the effect cannot be attributed to Luxembourgish supervision alone. Overall, the results show that multilingual fine-tuning including Luxembourgish can recover the performance lost in the EN+FR+DE setting, match the strongest single-language configuration overall, and provide the strongest Luxembourgish retrieval performance. Given the variability across three runs, the small differences between FT (FR) and EN+FR+DE+LB should be interpreted as comparable overall performance rather than a clear advantage for either setting.


\section{Conclusion}
Luxembourgish provides a representative low-resource use case for studying cross-lingual retrieval, given the scarcity of dedicated resources and retrieval benchmarks for the language. In this work, we introduced \textit{LëtzCross}, a cross-lingual benchmark for page-level retrieval over Luxembourgish PDF documents in a page-image RAG setting. The benchmark combines automatically generated text-focused QA pairs with manually authored visually grounded QA pairs. Our results show that late-interaction page-image retrievers outperform the evaluated OCR-based text-only baselines in a system-level comparison, and that fine-tuning improves retrieval with gains that transfer across query languages. Among the evaluated single-language settings, French fine-tuning yields the highest mean performance on Luxembourgish queries, while among the multilingual configurations, including Luxembourgish produces the strongest results and substantially improves Luxembourgish-query retrieval. Together, these findings provide a first benchmark and empirical study for cross-lingual page-level retrieval over Luxembourgish document pages.

\section*{Limitations}
This work has several limitations. First, \textit{LëtzCross} is relatively small, with a limited manually authored visually grounded subset compared with the automatically generated text-focused QA pairs. Consequently, the evaluation is dominated by text-focused queries, and results on the visually grounded subset should be considered complementary rather than representative of the benchmark. Second, the comparison between text-only and page-image retrievers is system-level: the text-only baselines rely on document parsing with OCR, layout analysis, and table extraction, while page-image retrievers operate on rendered pages. Alternative parsing or hybrid pipelines could therefore affect the observed gap. Third, all non-English queries are translated from English rather than independently authored, which may introduce translation artifacts or English-oriented formulations. Moreover, query translation and automatic QA generation rely on LLMs, although we apply validation and manual review to improve consistency and quality. Finally, our findings are specific to Luxembourgish and should be validated on other low-resource languages before broader generalization.

\section*{Acknowledgments}
This research was funded in whole or in part by the Luxembourg National Research Fund (FNR), grant reference NCER22/IS/16570468/NCER-FT. We acknowledge Google.org for providing Gemini credits used in this work. We also thank the anonymous reviewers for their careful reading and constructive feedback.

\bibliographystyle{acl_natbib}
\bibliography{latell}

\appendix

\end{document}